\documentclass{article}
\usepackage{ijcai22}

\usepackage{times}
\usepackage{soul}
\usepackage{url}
\usepackage[hidelinks]{hyperref}
\usepackage[utf8]{inputenc}
\usepackage[small]{caption}
\usepackage{graphicx}
\usepackage{amsmath}
\usepackage{amsthm}
\usepackage{booktabs}
\usepackage{algorithm}
\usepackage{algorithmic}

\makeatletter
\let\@internalcite\cite
     \def\cite{\def\citeauthoryear##1##2{##1, ##2}\@internalcite}
     \def\shortcite{\def\citeauthoryear##1{##2}\@internalcite}
     \def\@biblabel#1{\def\citeauthoryear##1##2{##1, ##2}[#1]\hfill}
\makeatother

\usepackage{amsmath,amssymb,amsfonts}
\usepackage{algorithmic}
\usepackage{graphicx}
\usepackage{textcomp}
\usepackage{multicol}
\usepackage{multirow}
\usepackage{comment}
\usepackage{tabularx}
\usepackage{xcolor, soul}

\usepackage{pifont}
\usepackage{multirow} % # robustness 표만들떄 필요

\makeatletter
\def\thickhline{%
  \noalign{\ifnum0=`}\fi\hrule \@height \thickarrayrulewidth \futurelet
   \reserved@a\@xthickhline}
\def\@xthickhline{\ifx\reserved@a\thickhline
               \vskip\doublerulesep
               \vskip-\thickarrayrulewidth
             \fi
      \ifnum0=`{\fi}}
\makeatother

\newlength{\thickarrayrulewidth}
\title{Global-Local Path Networks for Monocular Depth Estimation \\  with Vertical CutDepth}
\author{
Doyeon Kim$^1$\and
Woonghyun Ka$^2$\and
Pyunghwan Ahn$^1$\and
Donggyu Joo$^1$\and \\
Sewhan Chun$^2$\And
Junmo Kim$^{1,2}$
\affiliations
School of Electrical Engineering, KAIST, South Korea$^1$ \\
Division of Future Vehicle, KAIST, South Korea$^2$ 
\emails
\{doyeon\_kim, kwh950724, p.ahn, jdg105, alskdjfhgk, junmo.kim\}@kaist.ac.kr
}

\begin{document}

\maketitle

\begin{abstract}
    Depth estimation from a single image is an important task that can be applied to various fields in computer vision, and has grown rapidly with the development of convolutional neural networks. 
    In this paper, we propose a novel structure and training strategy for monocular depth estimation to further improve the prediction accuracy of the network. 
    We deploy a hierarchical transformer encoder to capture and convey the global context, and design a lightweight yet powerful decoder to generate an estimated depth map while considering local connectivity. 
    By constructing connected paths between multi-scale local features and the global decoding stream with our proposed selective feature fusion module, the network can integrate both representations and recover fine details. 
    In addition, the proposed decoder shows better performance than the previously proposed decoders,  with considerably less computational complexity. 
    Furthermore, we improve the depth-specific augmentation method by utilizing an important observation in depth estimation to enhance the model. 
    Our network achieves state-of-the-art performance over the challenging depth dataset NYU Depth V2. Extensive experiments have been conducted to validate and show the effectiveness of the proposed approach. Finally, our model shows better generalization ability and robustness than other comparative models. 
    The code will be available soon.
\end{abstract}

\section{Introduction}

Depth estimation is a challenging area that has been actively researched for many years. In particular, monocular depth estimation, which uses a single image to predict depth, is an ill-posed problem due to its inherent ambiguity. With the advent of convolutional neural networks (CNNs), many CNN-based approaches have been proposed for depth estimation and have yielded promising results~\cite{bhat2021adabins,lee2019big,fu2018deep}.
This paper proposes a new architecture and training strategy to further improve the performance by focusing on the essential properties of monocular depth estimation.

%First, we design a new global-local path network to fully extract and deliver  meaningful features in diverse scales. 
As many previous papers have claimed~\cite{chen2019structure,kim2020leveraging}, understanding both global and local contexts is crucial for successful depth estimation. There are many cues in monocular depth estimation that require understanding the scene on a global scale, such as the location of objects or the vanishing point. 
%However, due to the local behavior of convolutional layer, the network need to process further deeper layers to obtain large receptive field at low resolution. 
%Previous works mitigate this problem with Atrous Spatial Pooling (ASPP) or 
%Since \cite{eigen2014depth} has employed CNN for monocular depth estimation, many previous papers used CNN to tackle this problem. 
In addition, local connectivity of features is important because adjacent pixels tend to have similar values owing to their coplanar surfaces. %since they have coplanar surfaces. 
Therefore, we propose a new global-local path network to fully extract meaningful features on diverse scales and effectively deliver them throughout the network.
%Depth estimation is important not only for understanding a local area, but also for understanding the whole scene.
%Second, we found that the majority of the depth training dataset posses imbalanced distribution of the depth value. Due to the data acquisition process in depth estimation, it is natural to have imbalanced distribution in depth dataset, but it has been rarely explored. Consequently, it leads us to delve into more adaptive loss which takes care of the imbalance issue of training dataset.
%To achieve this, we facilitate the network to learn and convey global contexts with hierarchical transformer encoder. Recently, it is observed that transformer has the similar effect with enlarge the size of receptive field~\cite{}. Therefore, we construct the path for global context with encoder. 
First, we adopt a hierarchical transformer as the encoder to model long-range dependencies and capture multi-scale context features. 
%From 
In prior studies, it is observed that the transformer enables the network to enlarge the size of the receptive field~\cite{xie2021segformer}.
% Motivated by this knowledge, we leverage global relationships explicitly by building global path with multiple transformer blocks.
Motivated by this knowledge, we leverage the global relationships explicitly by building the global path with multiple transformer blocks.
% which consists of self-attention layers. 
%To leverage global relationship explicitly,  input image is fed through the multiple transformer blocks which consists of self-attention layers and build global path. 
%In the encoder path, input image is fed through the multiple transformer blocks to model long-range dependencies and capture multi-scale context features.
%Second, we design a highly utilized decoder with effective fusion module to make the path for local features and produce a fine depth map while preserving structural details.
Second, we design a highly utilized decoder with an effective fusion module to enable local features to produce a fine depth map while preserving structural details.
Contrary to the transformer, skip connections tend to create smaller receptive fields and help to focus on short-distance information~\cite{luo2016understanding}. 
%Thus, we take advantage of transformer and skip connection complementary by aggregating decoded and encoded features with our suggested fusion module.
Thus, the proposed architecture is intended to take the complementary advantages of both transformer and skip connections. This is enabled by aggregating the encoded and the decoded features using an input-dependent fusion module, called selective feature fusion (SFF).
%We name it selective feature fusion (SFF) since this module aid the model to selectively focus on the salient regions by estimating attention map for both features with the very fewer computational burden. 
The SFF module aids the model to selectively focus on the salient regions by estimating the attention map for both features with a very low computational burden.
Compared to other decoders, our decoder achieves superior performance with much lower complexity. 
%It supports that our decoder successfully fuse extracted feature and recover via global and local paths through entire framework.

%render representation.
%Many previous works adopted skip connection. but it is different from them that our model has highly less size and achieves better performance. 
%We design highly utilized decoder which can achieve better performance with only fewer convolution and bilinear upsampling layers. 
%With given obtained bottleneck feature through the encoder, our proposed lightweight decoder produces a fine depth map while preserving structural details. 
%Most of previous works conventionally use multiple bilinear upsampling with convolution or deconvolution layers. 
%However, we observe that the model can achieve better performance with only fewer convolution and bilinear upsampling layers.   

\begin{figure*}[h]
\centering
{\includegraphics[width=0.9\textwidth]{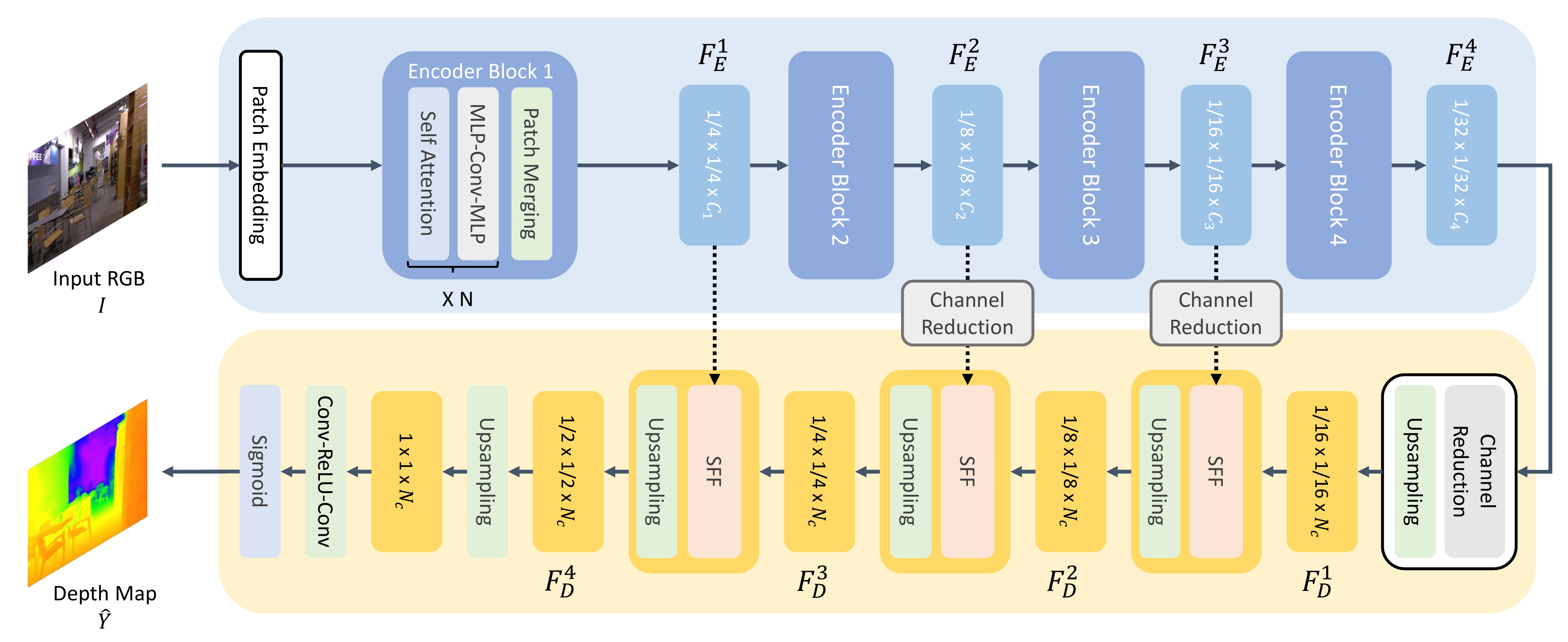}}
\caption{Overall architecture of the proposed network. The main components of the architecture are the encoder, decoder, and skip connections with feature fusion modules.}
\label{fig:architecture}
\end{figure*}

Furthermore, we train the network with an additional task specific data augmentation technique to boost the model capability. 
Data augmentation plays an important role in optimizing the network and can accelerate the model performance without additional computational costs. Nevertheless, data augmentation for depth estimation has been rarely adopted unlike for other tasks. 
To the best of our knowledge, \textit{CutDepth}~\cite{ishii2021cutdepth} is the first attempted data augmentation method specifically for depth estimation. %which is recently studied. %By attaching random parts from depth into image, the network is able to focus on the high-frequency regions of the RGB image which results in increasing capability of the model. 
We revisit CutDepth with the discovery that the vertical position of an object plays an essential role in monocular depth estimation~\cite{dijk2019neural}.
%Applying the suggested vertical CutDepth method, the model adaptively learns to capture vertical long-range information and results in improved performance.
To this end, we propose a variant of CutDepth, in which the crop is only applied to the horizontal axis, so that the model adaptively learns to capture vertical long-range information from the training sample.

The proposed network architecture and training strategy are experimented over the popular depth estimation dataset NYU Depth V2~\cite{silberman2012indoor} and exhibit the state-of-the-art performance. We validate the model through extensive quantitative and qualitative experiments, and the suggested architecture and data augmentation method demonstrate their effectiveness.
In addition, we observe that our network can generalize well under cross-dataset validation and shows robust performance against image corruption.

To summarize, our contributions are as follows:

\begin{itemize}
    \item We propose a novel global-local path architecture for monocular depth estimation.
    \item We suggest an improved depth-specific data augmentation method to boost the performance of the model.
    \item Our network achieves state-of-the-art performance on the most popular dataset NYU Depth V2 and shows higher generalization ability and robustness than previously developed networks.
\end{itemize}

\section{Related Work}
\noindent\textbf{Monocular depth estimation} is a computer vision task that predicts corresponding depth maps with given input images. 
Learning-based monocular depth estimation has been studied following the seminal work of \cite{saxena2008make3d} which used monocular cues to predict depth based on a Markov random field.
Later, with the development of CNNs, depth estimation networks have utilized the encoded features of deep CNNs that generalize well to various tasks and achieve drastic performance improvement~\cite{eigen2014depth,huynh2020guiding,yin2019enforcing}.
%Eigen~\etal \cite{eigen2015predicting, eigen2014depth} propose multi-scale networks that outputs coarse depth predictions at smaller scales, which were then refined via either lower-level features or an independent network. 
Recently, BTS~\cite{lee2019big} has suggested a local planar guidance layer that outputs plane coefficients, and then used them in the full resolution depth estimation. AdaBins~\cite{bhat2021adabins} reformulates the depth estimation problem as a classification task by dividing depth values into bins and shows state-of-the-art performance. 

\noindent\textbf{Transformer}~\cite{vaswani2017attention} adopts a self-attention mechanism with multi-layer perceptron (MLP) to overcome the limitation of previous RNN for natural language processing.
Since the emergence of the transformer, it has gained considerable attention in various fields. In the field of computer vision, a vision transformer (ViT)~\cite{dosovitskiy2020image} first uses a transformer to solve image classification tasks. 
%Despite using all the structures except for the embedding process, it attains excellent results compared to state-of-the-art convolutional networks. 
The success of ViT in the image classification task accelerates the introduction of the transformer into other tasks. 
SETR~\cite{zheng2021rethinking} first employs ViT as a backbone and demonstrates the potential of the transformer in dense prediction tasks by achieving new state-of-the-art performance.
\cite{xie2021segformer} proposed SegFormer, which is a transformer-based segmentation framework, with a simple lightweight MLP decoder.

%However, the use of transformer is rarely explored in monocular depth estimation compared to the other area. 
However, very few attempts have been made to employ a transformer for monocular depth estimation. Adabins~\cite{bhat2021adabins} uses a minimized version of a vision transformer (mini-ViT) to calculate bin width in an adaptive manner. 
DPT~\cite{ranftl2021vision} employs ViT as an encoder to obtain a global receptive field at different stages and attaches a convolutional decoder to make a dense prediction. 
However, both Adabins and DPT use CNN-based encoders and transformers simultaneously which increases the computational complexity. In addition, DPT is trained with an extra large-scale dataset. In contrast to these studies, our method use only one encoder and does not require an additional dataset to accomplish state-of-the-art performance.% compared to previous methods.
%But it requires massive amount of datasets to obtain competitive results with state-of-the-art works

%\noindent\textbf{Data imbalance} problem has received considerable attention in recent years, but it has mainly focused on classification task.
\noindent\textbf{Data augmentation} plays an important role in preventing overfitting by increasing the effective amount of training data. Therefore, common methods, such as flipping, color space transformation, cropping, and rotation, are used in several tasks to improve the network performance. However, although various methods, such as  CutMix~\cite{DBLP:journals/corr/abs-1905-04899}, Copy-Paste~\cite{ghiasi2021simple} and CutBlur~\cite{yoo2020rethinking}, have been actively proposed in diverse tasks, the depth-specific data augmentation method has rarely been studied. To the best of our knowledge, CutDepth~\cite{ishii2021cutdepth} is the first approach that attempts to augment the data in depth estimation. We accelerate the performance of this depth-specific data augmentation method by emphasizing the vertical location in the image.
%However, it is difficult to say that it considered the characteristic of depth estimation task enough because it is almost similar to ~\cite{DBLP:journals/corr/abs-1905-04899} except for using depth map instead of the raw image.

\section{Methods}

\subsection{Global-Local Path Networks}
Our depth estimation framework aims to predict the depth map $\hat{Y} \in \mathbb{R}^{H \times W \times 1}$ with a given RGB image $I \in \mathbb{R}^{H \times W \times 3}$. 
%As mentioned earlier, it is essential to exploit global and local contexts for scene understanding. Thus, 
Thus, we suggest a new architecture with global and local feature paths through the entire network to generate $\hat{Y}$. The overall structure of our framework is depicted in Figure~\ref{fig:architecture}. 
Our transformer encoder~\cite{xie2021segformer} enables the model to learn global dependencies, and the proposed decoder successfully recovers the extracted feature into the target depth map by constructing the local path through skip connection and the feature fusion module. 
We detail the proposed architecture in the following subsections. 

%\noindent\textbf{Encoder.} 
\subsection{Encoder}
In the encoding phase, we aim to leverage rich global information from an RGB image. 
To achieve this, we adopt a hierarchical transformer as the encoder.
%to consider semantic interdependency and exploit multi-scale context intermediate features. 
First, the input image $I$ is embedded as a sequence of patches with a $3 \times 3$ convolution operation. 
Then, the embedded patches are used as an input of the transformer block, which comprises of multiple sets of self-attention and the MLP-Conv-MLP layer with a residual skip. 
To reduce the computational cost in the self-attention layer, the dimension of each attention head is reduced with ratio $R_i$ for each $i$th block. 
%Then obtained feature transformed through MLP-$3 \times 3$ convolution-MLP layer with the residual skip.
With a given output, we perform patch merging with overlapped convolution. 
This process allows us to generate multi-scale features during the encoding phase and can be utilized in the decoding phase. 
We use four transformer blocks and each block generates $\frac{1}{4}$, $\frac{1}{8}$, $\frac{1}{16}$, $\frac{1}{32}$ scale feature with $[C_1, C_2, C_3, C_4]$ dimensions. %In this case, the value of $R_i$ and $C_i$ are $[8, 4, 2, 1]$ and $[64, 128, 320, 512]$.

%\noindent\textbf{Decoder.} 
\subsection{Lightweight Decoder} The encoder transforms the input image $I$ into the bottleneck feature $F_E^4$ with the size of $\frac{1}{32}H \times \frac{1}{32}W \times C_4$. To obtain the estimated depth map, we construct a lightweight and effective decoder to restore the bottleneck feature into the size of $H \times W \times 1$. 
Most of the previous studies have conventionally stacked multiple bilinear upsampling with convolution or deconvolution layers to recover the original size. 
However, we empirically observe that the model can achieve better performance with much fewer convolution and bilinear upsampling layers of the decoder if we design our restoring path effectively. 
First, we reduce the channel dimension of the bottleneck feature into $N_C$ with $1 \times 1$ convolution to avoid computational complexity. 
Then we use consecutive bilinear upsampling to enlarge the feature into size of $H \times W \times N_C$. 
Finally, the output is passed through two convolution layers and a sigmoid function to predict depth map $H \times W \times 1$. 
And depth map is multiplied with the maximum depth value to scale in meter. 
%To  restore $F_B$ into original size, consecutive decoder blocks with upsampling layer are applied until it reaches $H x W x 1$.
This simple decoder can generate as precise a depth map as other baseline structures.
%However, there still needs to exploit lower level to capture local structures with fine details.
However, to further exploit the local structures with fine details, we add skip connection with the proposed fusion module.

\subsection{Selective Feature Fusion} 
\begin{figure}[t]
\centering
{\includegraphics[width=0.5\textwidth]{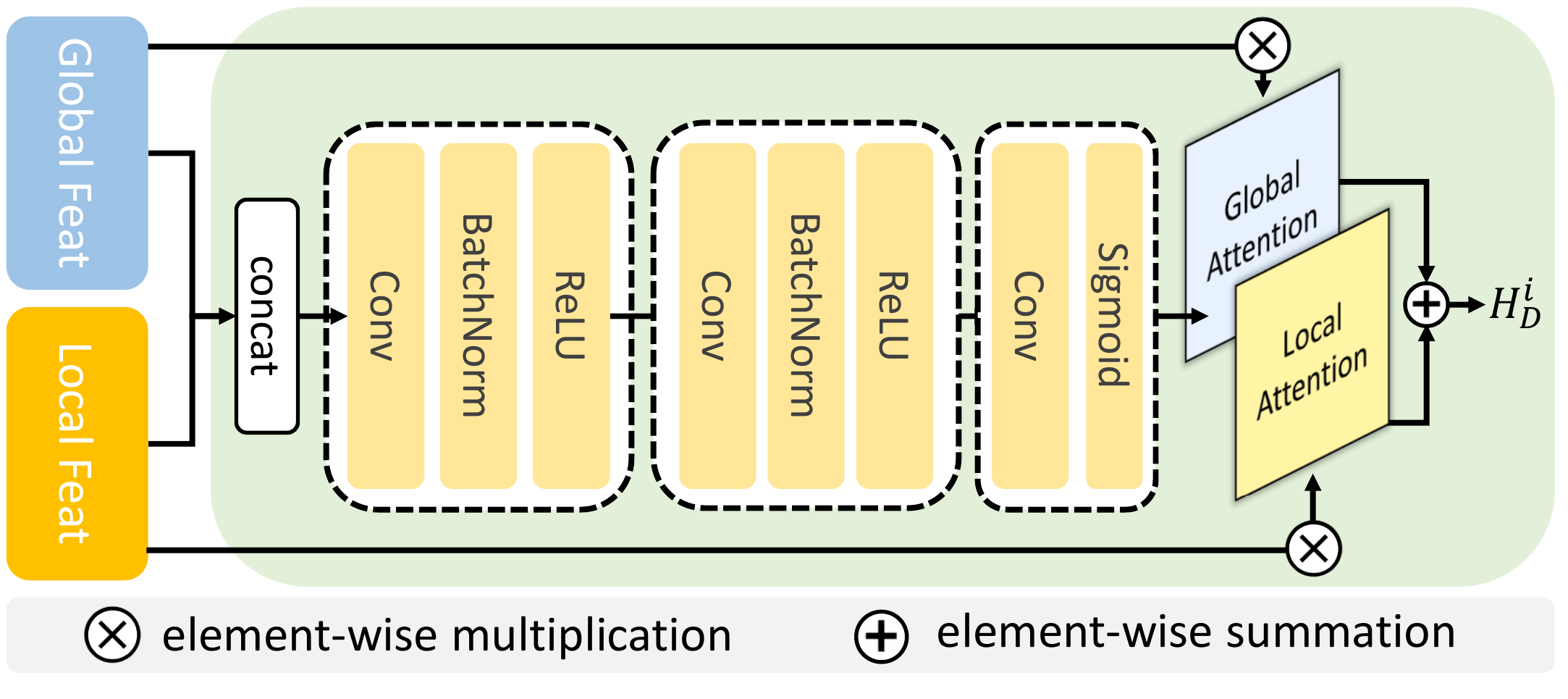}}
\caption{Detailed description of the SFF module.}
\label{fig:aff}
\end{figure}

We propose a Selective Feature Fusion (SFF) module to adaptively select and integrate local and global features by attaining an attention map for each feature. The detailed structure of SFF is illustrated in Figure~\ref{fig:aff}.
%We first reduce the dimension of multi-scale local context features to $N_C$ with convolution layer and concatenate with the decoded feature in channel dimension.
To match the dimensions of the decoded features $F_D$ and $F_E$, we first reduce the dimensions of multi-scale local context features to $N_C$ with the convolution layer.
Then, these features are concatenated along the channel dimension and passed through two $3 \times 3$ Conv-batch normalization-ReLU layers.
The final convolution and sigmoid layers produce a two-channel attention map, where each local and global feature is multiplied with each channel to focus on the significant location. Then these multiplied features are added element-wise to construct a hybrid feature  $H_D$.
To strengthen the local continuity we do not reduce the dimension on the $\frac{1}{4}$ scale feature. We will verify the effectiveness of the proposed decoder in section~\ref{sec:ablation}. 
%We will verify the choice of design and hyper-parameters in section~\ref{sec:ablation}. 
%In the Section~\ref{}, we show the effectiveness of our decoder. 

\begin{table*}[t]
 \centering
  \begin{tabular}{l | c| ccccccc}
    \toprule
        Method & Params (M)& $\delta_1 \uparrow$ & $\delta_2 \uparrow$ & $\delta_3 \uparrow$ & AbsRel $\downarrow$ & RMSE $\downarrow$ &  $ log_{10} \downarrow$ \\
        \midrule
        Eigen~\cite{eigen2014depth}  & 141 & 0.769 & 0.950 & 0.988 & 0.158 & 0.641 & -\\
        Fu~\cite{fu2018deep} & 110 & 0.828 & 0.965 & 0.992 & 0.115 & 0.509 & 0.051  \\
        Yin~\cite{yin2019enforcing} & 114 & 0.875 & 0.976 & 0.994 & 0.108 & 0.416 & 0.048  \\
        DAV~\cite{huynh2020guiding} & 25 & 0.882 & 0.980 & 0.996 & 0.108 & 0.412 & -  \\\textbf{}
        BTS~\cite{lee2019big} \hspace{1mm} & 47 & 0.885 & 0.978 & 0.994 & 0.110 & 0.392 & 0.047 \\
        Adabins\cite{bhat2021adabins} \hspace{1mm} & 78 & 0.903 & 0.984 & 0.997 & \underline{0.103} &  0.364 & \underline{0.044} \\
        DPT*~\cite{ranftl2021vision} \hspace{1mm} & 123 &  \underline{0.904} & \textbf{0.988} & \textbf{0.998} & 0.110 & \underline{0.357} & 0.045 \\
        \midrule
        \textbf{Ours} &  62 & \textbf{0.915} & \textbf{0.988} & \underline{0.997} & \textbf{0.098} & \textbf{0.344} & \textbf{0.042}\\
        %\hline
        \bottomrule
  \end{tabular}
  \vspace{-2mm}
  \caption{Performance on the NYU Depth V2 dataset. DPT* is trained with an extra dataset.}
  \label{tab:nyu}
\end{table*}

\iffalse
\begin{table}[t]
\centering
\resizebox{.5\textwidth}{!}{
\begin{tabular}{l | cccccc}
\toprule
Method    & $\delta_1 \uparrow$ & $\delta_2 \uparrow$ & $\delta_3 \uparrow$ & AbsRel $\downarrow$ & RMSE $\downarrow$ & log10 $\downarrow$ \\ 
\midrule
%Eigen~\etal~\cite{eigen2014depth} & 0.36 & 0.65 & 0.84 & 0.32 & 1.55 & 0.17 \\
%Liu~\etal~\cite{liu2018planenet} & 0.41 & 0.70 & 0.86 & 0.29 & 1.45 & 0.17 \\
Eigen~\etal~\cite{eigen2015predicting} & 0.47 & 0.78 & 0.93 & 0.25 & 1.26 & 0.13 \\
%Liu~\etal~\cite{liu2015deep} & 0.48 & 0.78 & 0.91 & 0.30 & 1.26 & 0.13 \\
%Laina~\etal~\cite{laina2016deeper} & 0.50 & 0.78 & 0.90 & 0.26 & 1.20 & 0.13 \\
VNL~\cite{yin2019enforcing} & 0.54 & 0.84 & 0.93 & 0.24 & 1.06 & 0.11 \\
BTS~\cite{lee2019big} & 0.54 & 0.86 & 0.95 & 0.23 & 0.93 & 0.11  \\
DORN~\cite{fu2018deep} & 0.55 & 0.81 & 0.92 & 0.24 & 1.13 & 0.12 \\
AdaBins~\cite{bhat2021adabins} & 0.55 & \underline{0.87} & \underline{0.96} & 0.21 & \underline{0.91} & 0.11 \\
%Li~\etal~\cite{li2017two} & 0.58 & 0.85 & 0.94 & 0.22 & 1.09 & 0.11 \\
SharpNet~\cite{ramamonjisoa2019sharpnet} & 0.59 & 0.84 & 0.94 & 0.26 & 1.07 & 0.11 \\
ACED~\cite{swami2020aced} & \underline{0.60} & \underline{0.87} & 0.95 & \underline{0.20} & 1.03 & \underline{0.10} \\
\midrule

Ours & \textbf{0.61} & \textbf{0.90} & \textbf{0.97} & \textbf{0.20} & \textbf{0.87} & \textbf{0.09} \\
\bottomrule
\end{tabular}
}

\caption{iBims-1 results. Methods are sorted using $\delta_1$.}
\label{tab:ibims1}
\end{table}
\fi

\subsection{Vertical CutDepth}

%Data augmentation enables the model to enhance the performance without influencing the model size or inference speed. 
%As mentioned earlier, data augmentation for depth estimation is rarely explored. 
Recently, a depth-specific data augmentation method named \textit{CutDepth} has been proposed~\cite{ishii2021cutdepth}, which replaces a part of the input image with the ground-truth depth map to provide diversity to the input image and enable the network to focus on the high-frequency area. 
In CutDepth, the coordinates $(l, u)$ and size $(w, h)$ of the cut region are randomly chosen. 
However, we believe that the vertical and horizontal directions should not be regarded equally for depth estimation based on the following discovery.
A previous study~\cite{dijk2019neural} suggested that the depth estimation networks mainly use vertical position in the image rather than apparent size or texture to predict the depth of arbitrary obstacles. 
This motivates us to propose \textit{vertical CutDepth}, which enhances the original CutDepth by preserving the vertical geometric information.
%It motivates us to propose Vertical CutDepth, a depth-specific augmentation method that preserves vertical geometric information.
%Let $x_{input}\in \mathbb{R}^{H \times W \times 3}$ be the input RGB image and $x_{depth}\in \mathbb{R}^{H \times W \times 1}$ be the corresponding depth map. 
%e 
%By replacing specific area on $I$ is by the same location of $Y$ along the vertical direction. 
%We replace chosen area on $I$ with the same location of $Y$ while not cropping along the vertical direction.
In vertical CutDepth, the ground-truth depth map replaces an area on $I$ with the same location of $Y$, but the crop is not applied along the vertical direction.
Therefore, the coordinate of the replacement region $(l, u)$ and size $(w, h)$ are calculated as follows:
%To determine the replacement region, horizontal coordinate $u$ and width $w$ are calculated as follows:
\iffalse
\begin{flalign}\label{eq1}
    %u &= 0 \\
    %l &= \alpha\times W \\
    (l, u) &= (\alpha\times W, 0) \\
    (w, h) &= (max((W-\alpha\times W)\times\beta\times p,1), H) 
    %w &= (max((W-\alpha\times W)\times\beta\times p,1) \\
\end{flalign}
\fi
\begin{equation}
\begin{split}
    (l, u) &= (\alpha\times W, 0) \\
    (w, h) &= (max((W-\alpha\times W)\times\beta\times p,1), H) 
\end{split}
\end{equation}
\noindent where $\alpha$ and $\beta$ are $\mathcal{U}(0,1)$. $p$ is a hyperparameter that is set at a value of $(0,1]$. By maintaining the vertical range of the input RGB image, the network can capture the long-range vertical direction for better prediction, as shown in the results. We set the value of $p$ to 0.75 by performing various settings of $p$ (Section~\ref{sec:ablation}).

\subsection{Training Loss}
In order to calculate the distance between predicted output $\hat{Y}$ and ground truth depth map $Y$,  we use scale-invariant log scale loss~\cite{eigen2014depth} to train the model. ${y_i}^*$ and $y_i$ indicates $i$th pixel in $\hat{Y}$ and $Y$. The equation of training loss is as follows:

\begin{equation}
    L = \frac{1}{n}\sum_i{{d_i}^2} - \frac{1}{2{n^2}}\left(\sum_i{d_i}^2\right)
\end{equation}
where $d_i = \log y_i - \log {y_i}^*$.

\section{Experiments}
To validate our approach, we perform several experiments on the NYU Depth V2 and SUN RGB-D datasets. We compare our model with existing methods through quantitative and qualitative evaluation, and an ablation study is conducted to show the effectiveness of each contribution. Additionally, we provide other results on additional dataset in supplementary material.

\subsection{Dataset}
\noindent\textbf{NYU Depth V2}
~\cite{silberman2012indoor} contains $640\times480$ images and corresponding depth maps of various indoor scenes acquired using a Microsoft Kinect camera. %The depth maps have an upper bound of 10 meters. 
We train our network using approximately 24K images on a random crop of $576\times448$ and test on 654 images. To facilitate a fair comparison, we perform the evaluation on a pre-defined center cropping by Eigen~\cite{eigen2014depth} with a maximum range of 10 m.

\noindent\textbf{SUN RGB-D}
~\cite{Song2015CVPR} contains approximately 10K RGB-D images of various indoor scenes captured by four different sensors, along with the corresponding depth and segmentation maps. We use this dataset for evaluating pre-trained models; thus, only the official test set of 5050 images is used.
% For inference, the input image is resized to $\frac{1}{32}$ of the original scale and then the prediction result is again resized to the original scale after inference.
Image sizes are not constant throughout this dataset, and thus we resize the image to the largest multiple of 32 below the image size, and then pass the resized image to predict the depth map, which is then resized to the original image.

\iffalse
\noindent\textbf{iBims-1}
Independent benchmark images and matched scans version 1~\cite{koch2018evaluation} (iBims-1) is a high-quality RGB-D dataset acquired using a digital single-lens reflex (DSLR) camera and high-precision laser scanner. 
%iBims-1 can be characterized by accurate edges and planar regions, consistent depth values, and accurate absolute distances. 
It contains 100 images for the test set and is only used to evaluate the methods. 
%which is considered to be a small number for training, and thus is generally used for testing of networks that are trained on other datasets.
\fi

\iffalse
\subsection{Evaluation Metrics}
To evaluate the performance of our method, we use standard metrics for depth estimation from prior works. In the following metrics, $d$ and $d^{*}$ respectively denotes predicted and ground-truth depth, and $D$ presents a set of all the predicted depth values of an input image.
\begin{itemize}
    \item $\delta_{t}$ : $\frac{1}{\left | D \right |} \left | \left \{ d\in D | max\left ( \frac{d^{*}}{d},\frac{d}{d^{*}} \right )<1.25^{t}\right \} \right |\times100\%$
    \item Abs Rel : $\frac{1}{\left | D \right |}\sum_{d\in D}\frac{\left | d^{*}-d \right |}{d^{*}}$
    \item Sq Rel : $\frac{1}{\left | D \right |}\sum_{d\in D}\frac{\left \| d^{*}-d \right \|^{2}}{d^{*}}$
    \item RMSE : $\sqrt{\frac{1}{\left | D \right |}\sum_{d\in D}\left \| d^{*}-d \right \|^{2}}$
    \item RMSE log : $\sqrt{\frac{1}{\left | D \right |}\sum_{d\in D}\left \| logd^{*}-logd \right \|^{2}}$
\end{itemize}
\fi

\subsection{Implementation Details}

We implement the proposed network using the PyTorch framework. For training, we use the one-cycle learning rate strategy with an Adam optimizer. %Adam optimizer~\cite{kingma2017adam} with $\beta_{1}=0.9$,$\beta_{2}=0.999$ and $\epsilon=10^{-8}$.
%We use the 1-cycle learning rate strategy. 
The learning rate increases from 3e-5 to 1e-4 following a poly LR schedule with a factor of 0.9 in the first half of the total iteration, and then decreases from 1e-4 to 3e-5 in the last half.
The total number of epochs is set to 25 with a batch size of 12. We use pre-trained weights from the  MiT-b4~\cite{xie2021segformer} network and initialize our encoder. The values of $N_C$, $R_i$ and $C_i$ are 64, $[8, 4, 2, 1]$ and $[64, 128, 320, 512]$, respectively.

In the case of data augmentation, the following strategies are used with the proposed vertical CutDepth with 50\% probability: horizontal flips, random brightness($\pm0.2$), contrast($\pm0.2$), gamma($\pm20$), hue($\pm20$), saturation($\pm30$), and value($\pm20$). %We also applied cutdepth(p=0.75) with 25\% probability and random crop of size $448\times576$. 
We apply $p=0.75$ for vertical CutDepth with 25\% possibility. %and random crop of size $448\times576$. 

\begin{table}[t]
\centering
\resizebox{.5\textwidth}{!}{
\begin{tabular}{l | cccccc}
\toprule
Method    & $\delta_1 \uparrow$ & $\delta_2 \uparrow$ & $\delta_3 \uparrow$ & AbsRel $\downarrow$ & RMSE $\downarrow$ & log10 $\downarrow$ \\ 
\midrule
Yin~\cite{yin2019enforcing} & 0.696 & 0.912 & 0.973 & 0.183 & 0.541 & 0.082 \\
BTS~\cite{lee2019big} & 0.740 & 0.933 & 0.980 & 0.172 & 0.515 & 0.075 \\
Adabins~\cite{bhat2021adabins} & \underline{0.771} & \underline{0.944} & \underline{0.983} & \underline{0.159} & \underline{0.476} & \underline{0.068} \\
\midrule
Ours & \textbf{0.814} & \textbf{0.964} & \textbf{0.991} & \textbf{0.144} & \textbf{0.418} & \textbf{0.061} \\
\bottomrule
\end{tabular}
}
\vspace{-2mm}
\caption{Performance on the SUN RGB-D dataset with the NYU Depth V2 trained model. We test the model without any fine-tuning.}
\label{tab:sun}
\end{table}

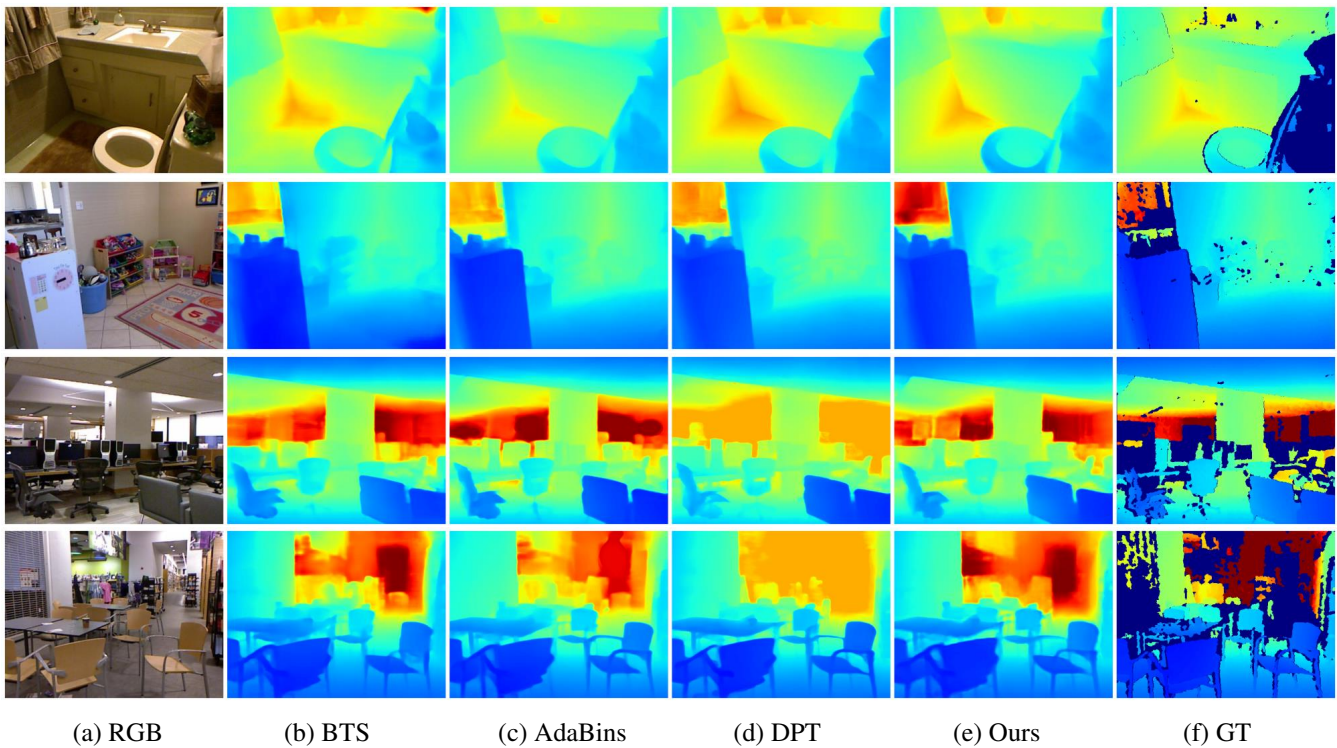
\begin{figure*}[h]
\centering
\begin{picture}(550,300)
\put(0,0){\includegraphics[width=\linewidth]{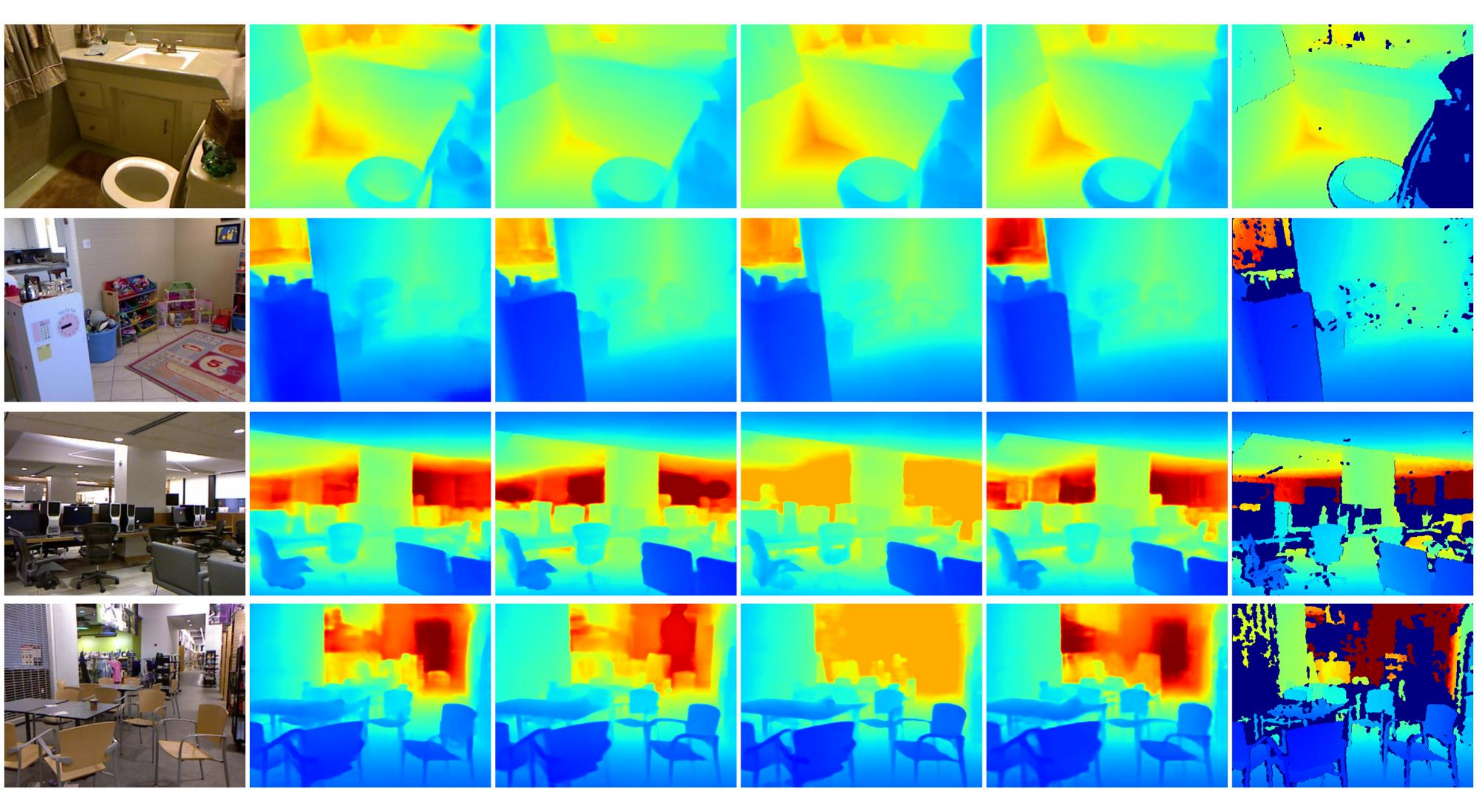}}
\put(27,-8){(a) RGB}
\put(107,-8){(b) BTS}
\put(188,-8){(c) AdaBins}
\put(277,-8){(d) DPT}
\put(360,-8){(e) Ours}
\put(447,-8){(f) GT}
\end{picture}
\vspace{0.5mm}
\caption{Qualitative comparison with previous works on the NYU Depth V2 dataset.}
\label{fig:nyu}
\end{figure*}

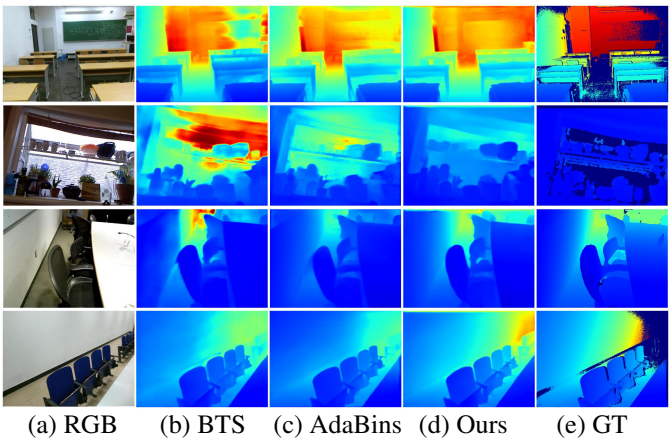
\begin{figure}[h]
\centering
%{\includegraphics[width=0.5\textwidth]{figures/qual_results_sun.pdf}}
\begin{picture}(390,150)
\put(0,0){\includegraphics[width=0.5\textwidth]{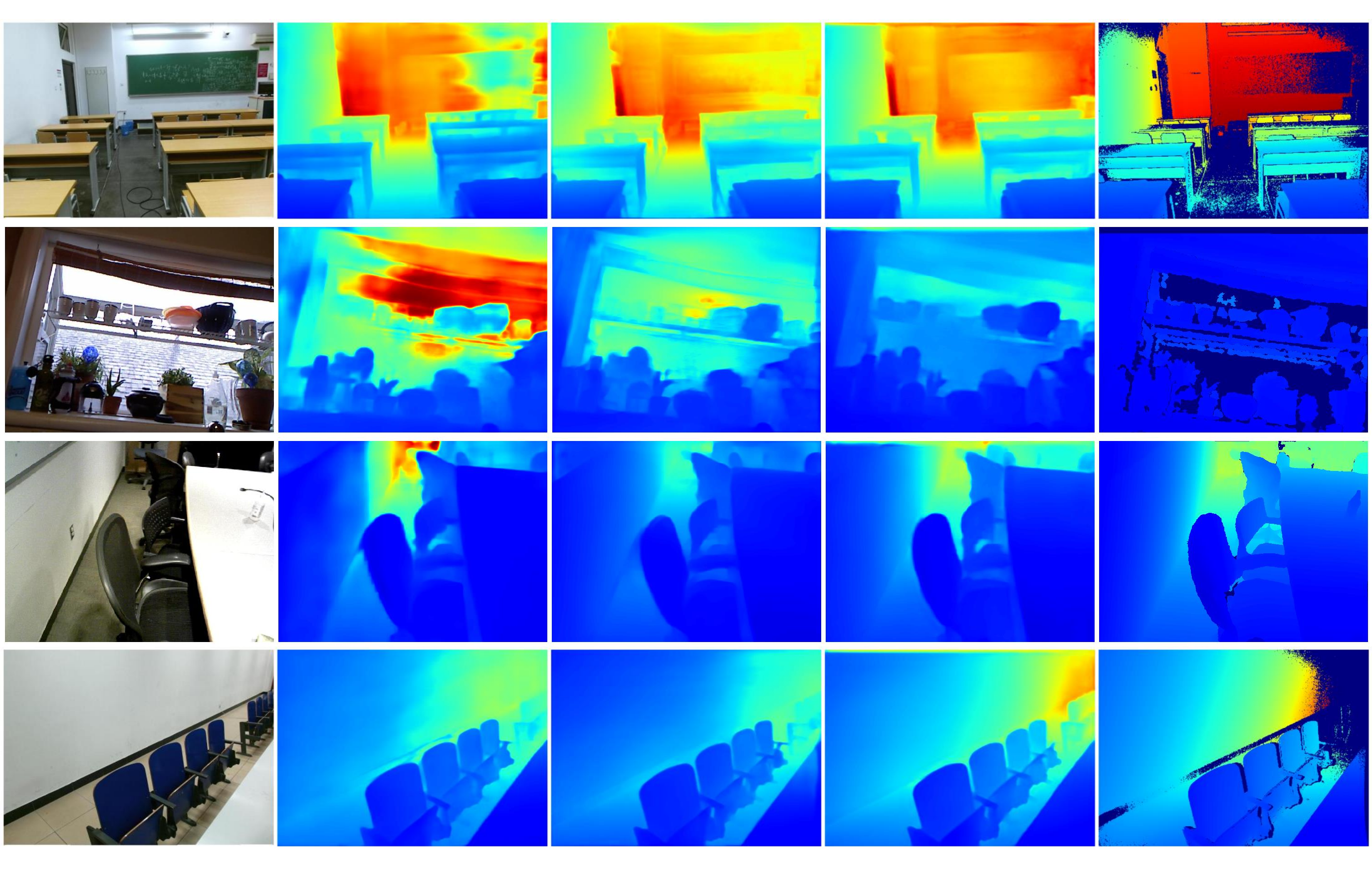}}
\put(10,-5){(a) RGB}
\put(60,-5){(b) BTS}
\put(103,-5){(c) AdaBins}
\put(158,-5){(d) Ours}
\put(210,-5){(e) GT}
\end{picture}
\caption{Examples of estimated depth maps on the SUN RGB-D dataset.}
\label{fig:sun_rgbd}
\end{figure}

\subsection{Comparison with State-of-the-Arts}

\noindent\textbf{NYU Depth V2.} Table~\ref{tab:nyu} presents the performance comparison of the NYU Depth V2 dataset.  DPT~\cite{ranftl2021vision} uses a much larger dataset of 1.4M images for training the model.  As listed in the table, the proposed model shows state-of-the-art performance in most of the evaluation metrics which we attribute to the proposed architecture and enhanced depth-specific data augmentation method. 
%Detailed contribution of each component is discussed in section~\ref{sec:ablation}.
%Our network shows the best performance in all metrics when compared with the other methods that used only NYU training set. 
%Both AdaBins and DPT are recently proposed methods that use transformer architectures as a component. 
Furthermore, our model achieves higher performance than the recently developed state-of-the-art models (Adabins, DPT) with lesser parameters. This suggests that the combination of the transformer encoder and the proposed compact decoder clearly makes an important contribution to estimate accurate depth maps in an efficient manner.
%Our model outperforms both methods in most metrics, which we attribute to the improved decoder architecture, data augmentation method, and the weighted loss function. Detailed contribution of each component is discussed in section~\ref{sec:ablation}. 
The visualized results are shown in Figure~\ref{fig:nyu}. In the figure, our model shows an accurate estimation of depth values for the provided example images and is more robust to various illumination conditions as compared to other methods. 

\noindent\textbf{SUN RGB-D.} We test our network on an additional indoor dataset SUN RGB-D to show the generalization performance. The network is trained on the NYU Depth V2 dataset and evaluated with a test set of SUN RGB-D without any fine-tuning process. Table~\ref{tab:sun} compares the results with those obtained by comparative studies. The proposed approach outperforms competing methods in all metrics. As shown in Figure~\ref{fig:sun_rgbd}, reasonable result depth maps are generated through our model without additional training.

\begin{table}[t]
\small
		\begin{center}
		\resizebox{.5\textwidth}{!}{
			\begin{tabular}{l | c | c c c c c c}
			    \toprule
				Method &  
				Params (M) &
				\begin{tabular}{@{}c@{}} $\delta_1 \uparrow$ \end{tabular}  & 
				\begin{tabular}{@{}c@{}} $\delta_2 \uparrow$ \end{tabular}  & 
				\begin{tabular}{@{}c@{}} $\delta_3 \uparrow$   \end{tabular} & 
				\begin{tabular}{@{}c@{}} AbsRel $\downarrow$  \end{tabular} &
				\begin{tabular}{@{}c@{}} RMSE $\downarrow$  \end{tabular} \\
				\midrule
				% check paramter numbers
				  Baseline - Dconv & 4.03 & 0.898 & 0.986 & \textbf{0.997} & 0.110 & 0.359\\
				  Baseline - UNet & 4.95 & 0.901 & \textbf{0.987} & \textbf{0.997} & 0.109 & 0.363 \\
				  \midrule
				  Ours (w/o SFF)  & \textbf{0.38} & 0.905 & 0.986 & \textbf{0.997} & 0.104 & 0.357 \\
				  %Ours (simple concat)  & &&&&& \\ 
				  \textbf{Ours}  & \textbf{0.66} & \textbf{0.908} & \textbf{0.987} & \textbf{0.997} & \textbf{0.101} & \textbf{0.351} \\
				  \midrule
				  \midrule
				  ~\cite{xie2021segformer} & 3.19 &  0.893 & 0.983 & 0.995 & 0.112  & 0.379 \\
				  ~\cite{lee2019big} & 5.79 &  0.906 & 0.985 & \textbf{0.997} & 0.102 & 0.356  \\
				  ~\cite{ranftl2021vision} & 14.15 & 0.907 & \textbf{0.987} & \textbf{0.997} & 0.103 & 0.354\\
				 \bottomrule
			\end{tabular}
			}
		\end{center}
		\vspace{-4mm}
		\caption{Comparison with multiple decoders. All results in this table are obtained from the same encoder.}
		\label{tab: decoder}
\end{table}

\begin{table*}[h]
\centering
\begin{tabular}{l | c | ccccccc}
\toprule
Corruption Type & Method & $\delta_1 \uparrow$ & $\delta_2 \uparrow$ & $\delta_3 \uparrow$ & AbsRel $\downarrow$ & SqRel $\downarrow$ & RMSE $\downarrow$ & RMSElog $\downarrow$ \\ \midrule
\multirow{3}{*}{Clean} & BTS & 0.885 & 0.978 & 0.994 & 0.110  & 0.066 & 0.392 & 0.142 \\
                       & Adabins & 0.903 & 0.984 & \textbf{0.997} & 0.103 & 0.057 & 0.364 & 0.130  \\
                       %& DPT* & 0.904 & 0.988 & 0.998 & 0.110  & 0.054 & 0.357 & 0.129 \\
                       & \textbf{Ours} & \textbf{0.915} & \textbf{0.988} & \textbf{0.997} & \textbf{0.098}  & \textbf{0.049} & \textbf{0.344} & \textbf{0.124} \\ \midrule
\multirow{3}{*}{Gaussian Noise} & BTS & 0.223 & 0.384 & 0.543 & 0.435 & 0.824 & 1.589 & 0.743 \\
                                & Adabins & 0.347 & 0.553 & 0.708 & 0.343 & 0.578 & 1.299 & 0.544 \\
                                %& DPT* & 0.760 & 0.936 & 0.983 & 0.152  & \textbf{0.124} & 0.571 & \textbf{0.197} \\
                                & \textbf{Ours} & \textbf{0.775} & \textbf{0.940} & \textbf{0.983} & \textbf{0.161}  & \textbf{0.126} & \textbf{0.541} & \textbf{0.198} \\ \midrule
\multirow{3}{*}{Motion Blur} & BTS & 0.677 & 0.850 & 0.922 & 0.189  & 0.207 & 0.701 & 0.279 \\
                             & Adabins & 0.697 & 0.859 & 0.927 & 0.180 & 0.182 & 0.643 & 0.262 \\
                             %& DPT* & 0.843 & 0.968 & 0.992 & 0.128  & 0.083 & 0.451 & 0.161 \\
                             & \textbf{Ours} & \textbf{0.807} & \textbf{0.946} & \textbf{0.981} & \textbf{0.139}  & \textbf{0.103} & \textbf{0.494} & \textbf{0.183} \\ \midrule
\multirow{3}{*}{Contrast} & BTS & 0.697 & 0.864 & 0.932 & 0.181  & 0.198 & 0.689 & 0.263 \\
                          & Adabins & 0.654 & 0.836 & 0.917 & 0.198  & 0.234 & 0.752 & 0.283 \\
                          %& DPT* & 0.853 & 0.973 & 0.994 & 0.121 & 0.075 & 0.436 & 0.152 \\
                          & \textbf{Ours} & \textbf{0.860} & \textbf{0.971} & \textbf{0.992} & \textbf{0.117}  & \textbf{0.074} & \textbf{0.427} & \textbf{0.152} \\ \midrule
\multirow{3}{*}{Snow} & BTS & 0.410 & 0.649 & 0.803 & 0.298 & 0.423 & 1.114 & 0.458 \\
                      & Adabins & 0.410 & 0.656 & 0.817 & 0.292 & 0.410 & 1.094 & 0.440 \\
                      %& DPT* & 0.619 & 0.881 & 0.968 & 0.205 & 0.196 & 0.740 & 0.261 \\
                      & \textbf{Ours} & \textbf{0.723} & \textbf{0.926} & \textbf{0.981} & \textbf{0.170} & \textbf{0.138} & \textbf{0.598} & \textbf{0.217} \\ \midrule
\end{tabular}

\vspace{-3mm}
\caption{Robustness experiment results on corrupted images of NYU Depth V2 datasets. The results of BTS and Adabins are obtained from distributed pre-trained weights.}
\label{tab:robustness_new}
\end{table*}

\begin{table}[t]
\small
		\label{tab: result}
		\begin{center}
		\resizebox{.5\textwidth}{!}{
			\begin{tabular}{l | c c c c c c c}
			    \toprule
				Method &  
				\begin{tabular}{@{}c@{}} $\delta_1 \uparrow$ \end{tabular}  & 
				\begin{tabular}{@{}c@{}} $\delta_2 \uparrow$ \end{tabular}  & 
				\begin{tabular}{@{}c@{}} $\delta_3 \uparrow$   \end{tabular} & 
				\begin{tabular}{@{}c@{}} AbsRel $\downarrow$  \end{tabular} &
				\begin{tabular}{@{}c@{}} RMSE $\downarrow$  \end{tabular} &
				\begin{tabular}{@{}c@{}} $log10$ $\downarrow$  \end{tabular} &\\
				\midrule
				% check paramter numbers
				  Baseline  & 0.908 & 0.987 & 0.997 & 0.101 & 0.351 & 0.043 \\
				  + CutDepth & 0.909 & 0.986 & 0.997 & 0.102 & 0.348 & \textbf{0.042} \\
				  \midrule
				  + Ours (p=0.25) & 0.911 & \textbf{0.988}  & 0.997 & 0.102 & 0.354 & 0.043 \\
				  + Ours (p=0.50) & 0.911 & \textbf{0.988} & 0.997 & 0.100 &  0.348 & \textbf{0.042} \\
				  + \textbf{Ours (p=0.75)} & \textbf{0.915} & \textbf{0.988} & \textbf{0.998} & \textbf{0.098} & \textbf{0.343} & \textbf{0.042} \\
				  + Ours (p=1.00) & 0.910 & \textbf{0.988} & 0.997 & 0.101 & 0.351 & 0.043\\
				  
				 \bottomrule
				
			\end{tabular}
			}
		\end{center}
		\vspace{-4mm}
		\caption{Experimental results with data augmentation. }
		\label{tab: cutdepth}
\end{table}

\subsection{Ablation Study}
\label{sec:ablation}
%\noindent\textbf{Effectiveness of proposed decoder} 
In this subsection, we validate the effectiveness of our approach through several experiments conducted on the NYU Depth V2 dataset. 

\noindent\textbf{Comparison with different decoder designs.}
Table~\ref{tab: decoder} demonstrates the comparison results with different decoder design. In this experiment, vertical CutDepth is omitted to solely compare the effectiveness of the decoder.
As our study aims to avoid computationally demanding decoders, we construct simple baselines and compare them with ours. Baseline-Dconv consists of consecutive deconvolution-batch normalization-ReLU blocks to obtain the desired depth map. 
In addition, Baseline-UNet is an improved structure from Baseline-Dconv that has skip connections between the encoder and decoder.
As detailed in the table, our decoder achieves better performance than the baselines.
Even without an SFF, it already shows better performance than other decoders while having fewer parameters. The powerful encoding ability of our encoder and the effectively designed decoder enables the network to produce a finely detailed depth map.
In addition, our proposed SFF leverages additional performance of our model.
%Moreover, we conduct two other multi-scale feature integration decoders which have been employed in previous transformer based architecture. 

% We conduct other comparison experiments with previously developed decoders that have employed multi-scale feature integration.
We additionally provide comparison with existing decoder architectures which integrates multi-scale features, in the bottom part of Table~\ref{tab: decoder}.
Despite the compactness of the proposed decoder, our network outperforms other networks. Our decoder has only 0.66M parameters while the MLP-decoder~\cite{xie2021segformer}, BTS~\cite{lee2019big} and DPT~\cite{ranftl2021vision} have 3.19M, 5.79M and 14.15M parameters, respectively, and thus, are highly heavier than ours. 
%It claims that if we well design the restoring path with skip connection and fuse module, it enables us to record fine performance with very fewer parameters with our powerful.
This indicates that we have effectively designed the restoring path for our encoder, which enables the proposed model to record fine performance with very few parameters.

\iffalse
\noindent\textbf{Design choice of the decoder.} 
To analyze the influence of the channel dimension $C$ in the decoder, we train the network with varying numbers of the dimension. From Table~\ref{tab: dimension}, it is observed that $C=64$ results in the most competitive model with reasonable computational cost. As the channel dimension increases, the model capacity and the performance tends to increase.  
However, our goal is to make a lightweight and powerful decoder that can express the best performance with fewer parameters, so it was designed with c=64.
\fi

\noindent\textbf{Effectiveness of the vertical CutDepth.}
We perform an ablation study on the data augmentation method used to train the network. The results are shown in Table~\ref{tab: cutdepth}. The first row of the table represents the baseline, which is only trained with traditional data augmentation except for CutDepth, and the second row shows the result obtained from adopting the basic CutDepth method. Then, we apply the proposed vertical CutDepth with different choices of hyperparameter $p$.
% the different numbers of hyper-parameters $p$.
As detailed in the table, CutDepth helps the model to achieve slightly better performance than the baseline. However, by applying vertical CutDepth, the network shows further improvement. This proves that utilizing vertical features enhances accurate depth estimation as compared to the case of simply cropping the random area. In addition, the model achieves the best performance with a setting of $p=0.75$.

\subsection{Robustness of the model}
In this subsection, we demonstrate the robustness of the proposed method against natural image corruptions. %In various safety-critical applications, 
Model robustness for depth estimation is essential because real world images always have a high possibility of being corrupted to a certain degree. Under these circumstances, it is beneficial to design a robust model so that it can perform the given task without being critically corrupted.
Following the previous study on the robustness of CNNs~\cite{hendrycks2018benchmarking}, we test our model on images that are corrupted by 16 different methods.
%Gaussian noise, shot noise, impulse noise, speckle noise, motion blur, defocus blur, glass blur, Gaussian blur, brightness, contrast, saturation, JPEG compression, snow, spatter, fog, and frost. 
Each corruption is applied with five different intensities, and the performance is averaged over all test images and all five intensities.

Table~\ref{tab:robustness_new} presents the depth estimation results for the corrupted images of the NYU Depth V2 test set. Due to space constraints, we provide the complete table in the supplementary material and present results on a few corruption types in Table~\ref{tab:robustness_new}. 
%For autonomous driving scenarios as in KITTI, various weather conditions are always possible and thus model robustness against these changes is critical. 
The results show that our model is clearly more robust to all types of corruption than the compared models.
The experimental results indicate that our model shows stronger robustness and thus is more appropriate for safety-critical applications.
%, including the ones in the supplementary material

\section{Conclusion}

This paper proposes a new architecture for monocular depth estimation to deliver meaningful global and local features and generate a precisely estimated depth map. 
We further exploit the depth-specific data augmentation technique to improve the performance of the model by considering the knowledge that the use of vertical position is a crucial property of depth estimation. 
The proposed method shows improvement over state-of-the-art performance for the NYU Depth V2 dataset. Moreover, extensive experimental results demonstrate the effectiveness and generalization ability of our network.

\bibliographystyle{named}
\bibliography{ijcai22}

\onecolumn

\section{Appendix: Additional dataset results}

In this section, we provide additional results on KITTI~\cite{geiger2013vision} and iBims-1~\cite{koch2018evaluation} datasets. KITTI is an outdoor depth estimation dataset and iBims-1 is an indoor dataset. %We trained our model with training set of KITTI and test with maximum range of 80m. %For iBims-1, trained model with NYU Depth V2 dataset has evaluated without any fine-tuning. 

\subsection{KITTI}
KITTI~\cite{geiger2013vision} contains stereo camera images and corresponding 3D LiDAR scans of various driving scenes acquired by car mounted sensors. The size of RGB images is around $1224\times368$. We train our network using approximately 23K images on a random crop of $704\times352$ and test on 697 images. To compare our performance with previous works, we use the crop as defined by Garg~\cite{garg2016unsupervised} and a maximum value of 80m for evaluation.
The results on the KITTI dataset are shown in Table~\ref{tab:kitti}. As shown in the table, our model outperforms other previous studies. 

\begin{table}[h]
 \centering
 \resizebox{0.8\textwidth}{!}{
  \begin{tabular}{l | c| ccccccc}
    \toprule
        Method & Params (M)& $\delta_1 \uparrow$ & $\delta_2 \uparrow$ & $\delta_3 \uparrow$ & AbsRel $\downarrow$ & RMSE $\downarrow$ &   RMSE log $ \downarrow$ \\
        \midrule
        %Eigen~\etal~\cite{eigen2014depth}  & - & 0.702 & 0.898 & 0.967 & 0.203 & 6.307 & 0.282 \\
        Fu~\cite{fu2018deep} & 110 & 0.932 & 0.984 & 0.994 & 0.072 & 2.727 & 0.120  \\
        Yin~\cite{yin2019enforcing} & 114 & 0.938 & 0.984 & \underline{0.998} & 0.072 & 3.258 & 0.117  \\
        BTS~\cite{lee2019big} \hspace{1mm} & 113 & 0.956 & 0.993 & \underline{0.998} & 0.059 & 2.756 & 0.088 \\
        DPT*~\cite{ranftl2021vision} \hspace{1mm} & 123 &  0.959 & 0.995 & \textbf{0.999} & 0.062 & 2.573 & 0.092 \\
        Adabins~\cite{bhat2021adabins} \hspace{1mm} & 78 & \underline{0.964} & \underline{0.995} & \textbf{0.999} & \underline{0.058} &  \underline{2.360} & \underline{0.088} \\
        \midrule
        \textbf{Ours} &  62 & \textbf{0.967} & \textbf{0.996} & \textbf{0.999} & \textbf{0.057} & \textbf{2.297} & \textbf{0.086}\\
        %\hline
        \bottomrule
  \end{tabular}
  }
  \caption{Performance on the KITTI dataset. DPT* is trained with extra dataset.}
  \label{tab:kitti}
\end{table}

\subsection{iBims-1}
iBims-1~\cite{koch2018evaluation} (independent Benchmark images and matched scans version 1) is a high quality RGB-D dataset acquired using a digital single-lens reflex (DSLR) camera and high-precision laser scanner. iBims-1 can be characterized by accurate edges and planar regions, consistent depth values, and accurate absolute distances.
We evaluate with our NYU Depth V2 trained model without any fine-tuning. Results on iBims-1 dataset are listed in Table~\ref{tab:ibims1}.
%It contains 100 images, which is considered to be a small number for training, and thus is generally used for testing of networks that are trained on other datasets.

\begin{table}[h]
\centering
\resizebox{.8\textwidth}{!}{
\begin{tabular}{l | cccccc}
\toprule
Method    & $\delta_1 \uparrow$ & $\delta_2 \uparrow$ & $\delta_3 \uparrow$ & AbsRel $\downarrow$ & RMSE $\downarrow$ & log10 $\downarrow$ \\ 
\midrule
%Eigen~\etal~\cite{eigen2015predicting} & 0.47 & 0.78 & 0.93 & 0.25 & 1.26 & 0.13 \\
VNL~\cite{yin2019enforcing} & 0.54 & 0.84 & 0.93 & 0.24 & 1.06 &\underline{0.11} \\
% BTS~\cite{lee2019big} & 0.54 & 0.86 & 0.95 & 0.23 & 0.93 & 0.11  \\ % AdaBins crop
BTS~\cite{lee2019big} & 0.53 & 0.84 & 0.94 & 0.24 & 1.10 & 0.12  \\ % no crop
DORN~\cite{fu2018deep} & 0.55 & 0.81 & 0.92 & 0.24 & 1.13 & 0.12 \\
% AdaBins~\cite{bhat2021adabins} & 0.55 & \underline{0.87} & \underline{0.96} & 0.21 & \underline{0.91} & 0.11 \\ % AdaBins crop
AdaBins~\cite{bhat2021adabins} & 0.55 & 0.86 & \underline{0.95} & \underline{0.22} & 1.07 & \underline{0.11} \\ % no crop
SharpNet~\cite{ramamonjisoa2019sharpnet} & 0.59 & 0.84 & 0.94 & 0.26 & 1.07 & \underline{0.11} \\
ACED~\cite{swami2020aced} & \underline{0.60} & \underline{0.87} & \underline{0.95} & \textbf{0.20} & \underline{1.03} & \textbf{0.10} \\
\midrule
% Ours & \textbf{0.62} & \textbf{0.91} & \textbf{0.98} & \textbf{0.19} & \textbf{0.82} & \textbf{0.09} \\ % AdaBins crop
Ours & \textbf{0.61} & \textbf{0.89} & \textbf{0.96} & \textbf{0.20} & \textbf{1.01} & \textbf{0.10} \\ % no crop
\bottomrule
\end{tabular}
}

\caption{Performance on the iBims-1 dataset.}
\label{tab:ibims1}
\end{table}

\clearpage
\section{Appendix: Robustness of the Model}

In Table~\ref{tab:robustness_full}, we report a full table of the results on the corrupted NYU Depth V2 dataset. (Section 4.5 of the main paper)

\begin{table*}[!htbp]
\centering
\resizebox{0.8\textwidth}{!}{
\begin{tabular}{l | c | c | ccccccc}
\toprule
\multicolumn{2}{c|}{Corruption Type}
& Method & $\delta_1 \uparrow$ & $\delta_2 \uparrow$ & $\delta_3 \uparrow$ & AbsRel $\downarrow$ & SqRel $\downarrow$ & RMSE $\downarrow$ & RMSElog $\downarrow$ \\ \midrule
\multicolumn{2}{c|}{\multirow{3}{*}{Clean}} & BTS & 0.885 & 0.978 & 0.994 & 0.110  & 0.066 & 0.392 & 0.142 \\
                       \multicolumn{2}{c|}{} & Adabins & 0.903 & 0.984 & \textbf{0.997} & 0.103 & 0.057 & 0.364 & 0.130  \\
                       %& DPT* & 0.904 & 0.988 & 0.998 & 0.110  & 0.054 & 0.357 & 0.129 \\
                       \multicolumn{2}{c|}{} & \textbf{Ours} & \textbf{0.915} & \textbf{0.988} & \textbf{0.997} & \textbf{0.098}  & \textbf{0.049} & \textbf{0.344} & \textbf{0.124} \\ 
                       \midrule
\multirow{12}{*}{Noise} & \multirow{3}{*}{Gaussian Noise} & BTS & 0.223 & 0.384 & 0.543 & 0.435 & 0.824 & 1.589 & 0.743 \\
                                & & Adabins & 0.347 & 0.553 & 0.708 & 0.343 & 0.578 & 1.299 & 0.544 \\
                                %& DPT* & 0.760 & 0.936 & 0.983 & 0.152  & \textbf{0.124} & 0.571 & \textbf{0.197} \\
                                & & \textbf{Ours} & \textbf{0.775} & \textbf{0.940} & \textbf{0.983} & \textbf{0.161}  & \textbf{0.126} & \textbf{0.541} & \textbf{0.198} \\ \cmidrule{2-10}
      &  \multirow{3}{*}{Shot} & BTS & 0.280 & 0.448 & 0.600 & 0.399 & 0.736 & 1.482 & 0.669 \\
                                & & Adabins & 0.436 & 0.653 & 0.791 & 0.293 & 0.460 & 1.141 & 0.454 \\
                                %& DPT* & 0.760 & 0.936 & 0.983 & 0.152  & \textbf{0.124} & 0.571 & \textbf{0.197} \\
                                & & \textbf{Ours} & \textbf{0.791} & \textbf{0.949} & \textbf{0.986} & \textbf{0.152}  & \textbf{0.114} & \textbf{0.523} & \textbf{0.189} \\ \cmidrule{2-10}
    &  \multirow{3}{*}{Impulse} & BTS & 0.116 & 0.249 & 0.420 & 0.504 & 1.006 & 1.818 & 0.875 \\
                                & & Adabins & 0.377 & 0.589 & 0.736 & 0.327 & 0.541 & 1.246 & 0.518 \\
                                %& DPT* & 0.760 & 0.936 & 0.983 & 0.152  & \textbf{0.124} & 0.571 & \textbf{0.197} \\
                                & & \textbf{Ours} & \textbf{0.760} & \textbf{0.938} & \textbf{0.984} & \textbf{0.167}  & \textbf{0.131} & \textbf{0.556} & \textbf{0.204} \\ \cmidrule{2-10}
    &  \multirow{3}{*}{Speckle} & BTS & 0.456 & 0.633 & 0.756 & 0.302 & 0.500 & 1.159 & 0.492\\
                                & & Adabins & 0.639 & 0.834 & 0.918 & 0.200 & 0.244 & 0.805 & 0.294 \\
                                %& DPT* & 0.760 & 0.936 & 0.983 & 0.152  & \textbf{0.124} & 0.571 & \textbf{0.197} \\
                                & & \textbf{Ours} & \textbf{0.830} & \textbf{0.965} & \textbf{0.991} & \textbf{0.136}  & \textbf{0.091} & \textbf{0.467} & \textbf{0.168} \\ \midrule
\multirow{12}{*}{Blur} & \multirow{3}{*}{Motion} & BTS & 0.677 & 0.850 & 0.922 & 0.189 & 0.207 & 0.701 & 0.279 \\
                                & & Adabins & 0.697 & 0.859 & 0.927 & 0.180 & 0.182 & 0.643 & 0.262 \\
                                %& DPT* & 0.760 & 0.936 & 0.983 & 0.152  & \textbf{0.124} & 0.571 & \textbf{0.197} \\
                                & & \textbf{Ours} & \textbf{0.807} & \textbf{0.946} & \textbf{0.981} & \textbf{0.139}  & \textbf{0.103} & \textbf{0.494} & \textbf{0.183} \\ \cmidrule{2-10}
      &  \multirow{3}{*}{Defocus} & BTS & 0.511 & 0.684 & 0.786 & 0.276 & 0.415 & 1.002 & 0.436 \\
                                & & Adabins & 0.599 & 0.769 & 0.859 & 0.227 & 0.277 & 0.793 & 0.341 \\
                                %& DPT* & 0.760 & 0.936 & 0.983 & 0.152  & \textbf{0.124} & 0.571 & \textbf{0.197} \\
                                & & \textbf{Ours} & \textbf{0.728} & \textbf{0.897} & \textbf{0.954} & \textbf{0.166}  & \textbf{0.155} & \textbf{0.605} & \textbf{0.228} \\ \cmidrule{2-10}
    &  \multirow{3}{*}{Glass} & BTS & 0.671 & 0.855 & 0.927 & 0.193 & 0.224 & 0.747 & 0.285 \\
                                & & Adabins & 0.743 & \textbf{0.914} & 0.967 & 0.165 & 0.149 & 0.619 & 0.223 \\
                                %& DPT* & 0.760 & 0.936 & 0.983 & 0.152  & \textbf{0.124} & 0.571 & \textbf{0.197} \\
                                & & \textbf{Ours} & \textbf{0.770} & \textbf{0.914} & \textbf{0.978} & \textbf{0.155}  & \textbf{0.132} & \textbf{0.573} & \textbf{0.202} \\ \cmidrule{2-10}
    &  \multirow{3}{*}{Gaussian} & BTS & 0.530 & 0.688 & 0.779 & 0.274 & 0.422 & 0.989 & 0.437\\
                                & & Adabins & 0.595 & 0.738 & 0.814 & 0.244 & 0.341 & 0.847 & 0.379 \\
                                %& DPT* & 0.760 & 0.936 & 0.983 & 0.152  & \textbf{0.124} & 0.571 & \textbf{0.197} \\
                                & & \textbf{Ours} & \textbf{0.716} & \textbf{0.865} & \textbf{0.926} & \textbf{0.177} & \textbf{0.190} & \textbf{0.641} & \textbf{0.248} \\ \midrule
\multirow{12}{*}{Digital} & \multirow{3}{*}{Brightness} & BTS & 0.842 & 0.965 & 0.990 & 0.124 & 0.084 & 0.457 & 0.166  \\
                                & & Adabins & 0.862 & 0.972 & 0.994 & 0.117 & 0.073 & 0.427 & 0.152 \\
                                %& DPT* & 0.760 & 0.936 & 0.983 & 0.152  & \textbf{0.124} & 0.571 & \textbf{0.197} \\
                                & & \textbf{Ours} & \textbf{0.899} & \textbf{0.984} & \textbf{0.997} & \textbf{0.104} & \textbf{0.055} & \textbf{0.369} & \textbf{0.133} \\ \cmidrule{2-10}
      &  \multirow{3}{*}{Contrast} & BTS & 0.697 & 0.864 & 0.932 & 0.181 & 0.198 & 0.689 & 0.263 \\
                                & & Adabins & 0.654 & 0.836 & 0.917 & 0.198 & 0.234 & 0.752 & 0.283\\
                                %& DPT* & 0.760 & 0.936 & 0.983 & 0.152  & \textbf{0.124} & 0.571 & \textbf{0.197} \\
                                & & \textbf{Ours} & \textbf{0.860} & \textbf{0.971} & \textbf{0.992} & \textbf{0.117} & \textbf{0.074} & \textbf{0.427} & \textbf{0.152} \\ \cmidrule{2-10}
    &  \multirow{3}{*}{Saturation} & BTS & 0.814 & 0.950 & 0.983 & 0.135 & 0.103 & 0.505 & 0.182 \\
                                & & Adabins & 0.839 & 0.965 & 0.991 & 0.125 & 0.086 & 0.465 & 0.162 \\
                                %& DPT* & 0.760 & 0.936 & 0.983 & 0.152  & \textbf{0.124} & 0.571 & \textbf{0.197} \\
                                & & \textbf{Ours} & \textbf{0.896} & \textbf{0.984} & \textbf{0.996} & \textbf{0.107} & \textbf{0.058} & \textbf{0.374} & \textbf{0.134}\\ \cmidrule{2-10}
    &  \multirow{3}{*}{JPEG Compression} & BTS & 0.786 & 0.942 & 0.983 & 0.154 & 0.124 & 0.532 & 0.195 \\
                                & & Adabins & 0.804 & 0.954 & 0.988 & 0.153 & 0.115 & 0.493 & 0.182 \\
                                %& DPT* & 0.760 & 0.936 & 0.983 & 0.152  & \textbf{0.124} & 0.571 & \textbf{0.197} \\
                                & & \textbf{Ours} & \textbf{0.860} & \textbf{0.973} & \textbf{0.994} & \textbf{0.123} & \textbf{0.073} & \textbf{0.413} & \textbf{0.153} \\ \midrule                   
                            
\multirow{12}{*}{Weather} & \multirow{3}{*}{Snow} & BTS & 0.410 & 0.649 & 0.803 & 0.298 & 0.423 & 1.114 & 0.458 \\
                                & & Adabins & 0.410 & 0.656 & 0.817 & 0.292 & 0.410 & 1.094 & 0.440  \\
                                %& DPT* & 0.760 & 0.936 & 0.983 & 0.152  & \textbf{0.124} & 0.571 & \textbf{0.197} \\
                                & & \textbf{Ours} & \textbf{0.723} & \textbf{0.926} & \textbf{0.981} & \textbf{0.170} & \textbf{0.138} & \textbf{0.598} & \textbf{0.217} \\ \cmidrule{2-10}
      &  \multirow{3}{*}{Spatter} & BTS &0.705 & 0.878 & 0.945 & 0.176 & 0.168 & 0.642 & 0.250 \\
                                & & Adabins & 0.699 & 0.890 & 0.964 & 0.173 & 0.155 & 0.625 & 0.234 \\
                                %& DPT* & 0.760 & 0.936 & 0.983 & 0.152  & \textbf{0.124} & 0.571 & \textbf{0.197} \\
                                & & \textbf{Ours} & \textbf{0.835} & \textbf{0.971} & \textbf{0.994} & \textbf{0.134} & \textbf{0.083} & \textbf{0.445} & \textbf{0.162} \\ \cmidrule{2-10}
    &  \multirow{3}{*}{Fog} & BTS & 0.588 & 0.798 & 0.893 & 0.227 & 0.273 & 0.835 & 0.332 \\
                                & & Adabins & 0.523 & 0.748 & 0.873 & 0.252 & 0.308 & 0.898 & 0.357 \\
                                %& DPT* & 0.760 & 0.936 & 0.983 & 0.152  & \textbf{0.124} & 0.571 & \textbf{0.197} \\
                                & & \textbf{Ours} & \textbf{0.759} & \textbf{0.928} & \textbf{0.978} & \textbf{0.153} & \textbf{0.125} & \textbf{0.559} & \textbf{0.204} \\ \cmidrule{2-10}
    &  \multirow{3}{*}{Frost} & BTS & 0.515 & 0.734 & 0.850 & 0.261 & 0.359 & 0.996 & 0.400 \\
                                & & Adabins & 0.439 & 0.691 & 0.842 & 0.280 & 0.398 & 1.074 & 0.413 \\
                                %& DPT* & 0.760 & 0.936 & 0.983 & 0.152  & \textbf{0.124} & 0.571 & \textbf{0.197} \\
                                & & \textbf{Ours} & \textbf{0.736} & \textbf{0.929} & \textbf{0.983} & \textbf{0.163} & \textbf{0.130} & \textbf{0.576} & \textbf{0.209} \\ \midrule
\end{tabular}
}
\caption{Robustness experiment results on corrupted images of NYU Depth V2 datasets.}
\label{tab:robustness_full}
\end{table*}

\section{Appendix: Detailed structure of baseline decoder}

We illustrate the detailed structure of Baseline-DConv and Baseline-UNet in  Figure~\ref{fig:structure}. We use transposed convolution with $K=3, S=2, P=1$ parameters to upscale the given feature into 2x size. For Baseline-UNet, features from encoder $F_E^3, F_E^2, F_E^1$ are concatenated in channel dimension.

\begin{table*}[!ht]
  \begin{center}
  \centering
  \begin{tabular}{ cc }
    %\hline
    (a) &
    \begin{minipage}{\linewidth}
      \includegraphics[width=\linewidth]{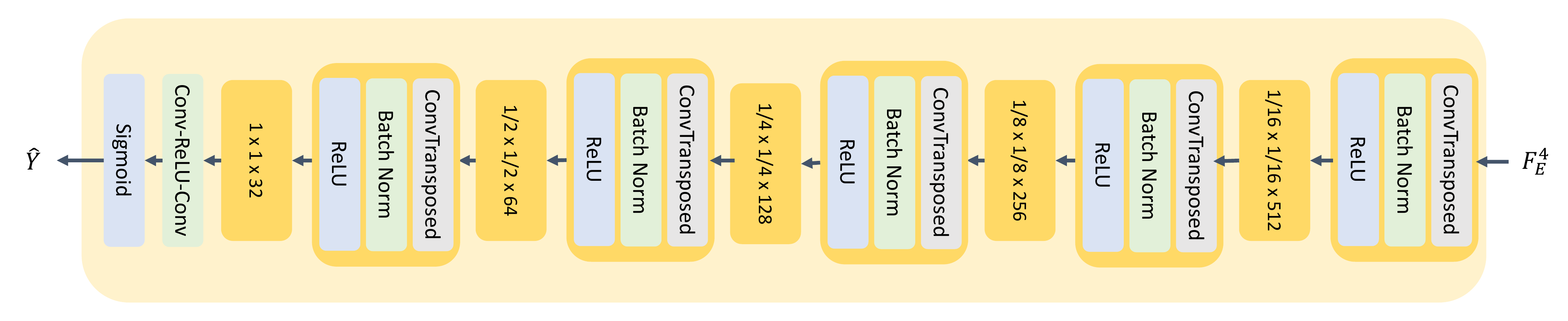}
    \end{minipage} \\

    \rule{0pt}{4ex}  (b) &
    \begin{minipage}{\linewidth}
      \includegraphics[width=\linewidth]{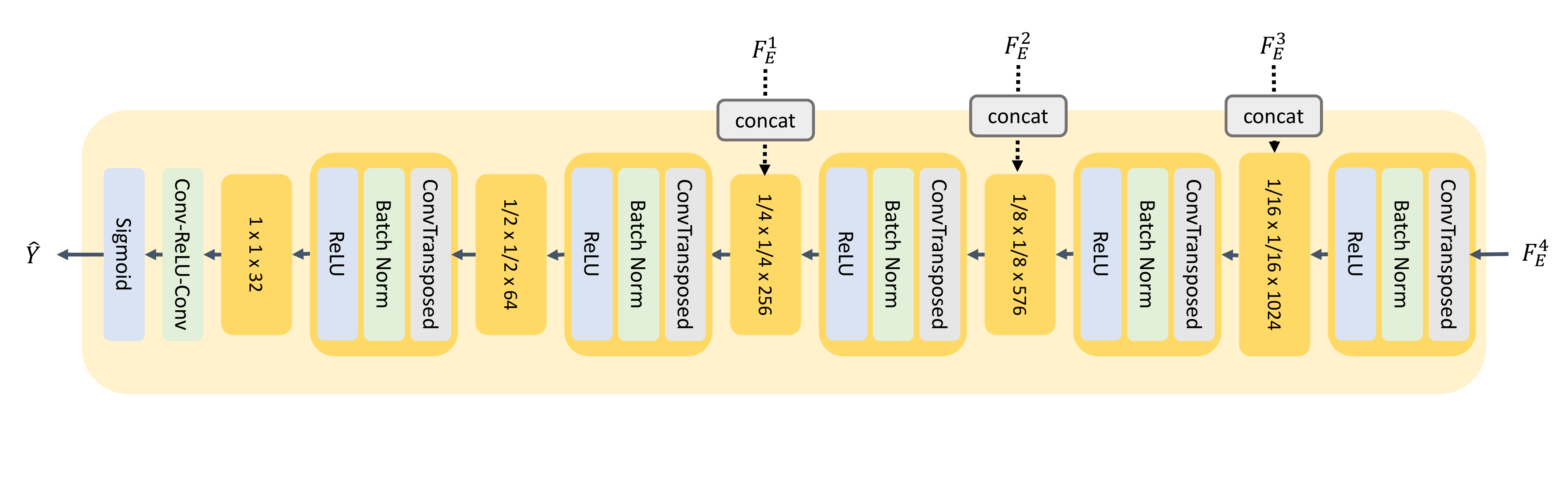}
    \end{minipage}
    
  \end{tabular}
  \end{center}
%\vspace{3mm}
\captionof{figure}{The detailed structure of (a) Baseline-DConv (b) Baseline-UNet. }
\label{fig:structure}
\end{table*}

\end{document}